\documentclass[accepted]{uai2022} 

\usepackage{mathtools,graphicx}
\usepackage{amsmath,amssymb, bm, bbm, amsthm}
\usepackage{subfig}
\usepackage{algpseudocode}
\usepackage{algorithm}
\newtheorem{definition}{Definition}
\newtheorem{proposition}{Proposition}
\newtheorem{corollary}{Corollary}
\newtheorem{lemma}{Lemma}
\theoremstyle{remark}
\newtheorem{remark}{Remark}

\DeclareMathOperator*{\argmin}{arg\,min}
\DeclareMathOperator{\E}{\mathbb{E}}

\DeclarePairedDelimiter\ceil{\lceil}{\rceil}
\title{Assessment of Prediction Intervals Using Uncertainty Characteristics Curves}

\author[1]{\href{mailto:<jiri@us.ibm.com>?Subject=Your UAI 2022 paper}{Ji\v{r}\'\i\, Navr\'atil}}
\author[1]{Benjamin Elder}
\author[1]{Matthew Arnold}
\author[1]{Soumya Ghosh}
\author[1]{Prasanna Sattigeri}
\affil[1]{%
    IBM Thomas J. Watson Research Center\\
	Yorktown Heights, NY 10598, USA
}

\begin{document}
%
\maketitle

\begin{abstract}
Accurate quantification of model uncertainty has long been recognized as a fundamental requirement for trusted AI. In regression tasks, 
uncertainty is typically quantified using prediction intervals calibrated to an ad-hoc operating point, making evaluation and comparison across different studies relatively difficult.
Our work leverages: (1) the concept of operating characteristics curves and (2) the notion of a gain over a null reference, to derive a novel operating point agnostic assessment methodology for prediction intervals.
The paper defines the Uncertainty Characteristics Curve and demonstrates its utility in selected scenarios. We argue that the proposed method addresses the current need for comprehensive assessment of prediction intervals and thus represents a valuable addition to the uncertainty quantification toolbox.   
\end{abstract}

\vspace{-.1cm}
\section{Introduction}\label{Sec:Intro}

The ability to quantify the uncertainty of a model is one of the fundamental requirements in trusted, safe, and actionable AI \cite{Jiang2018_trust, Arnold2019_factsheets, Begoli2019}. 
Numerous methods of generating uncertainty bounds (referred to as {\em prediction intervals}, or error bounds) 
have been proposed in statistics and machine learning literature. 

Evaluating the quality of prediction intervals (PI), however, remains challenging. While metrics such as the likelihood are popular, they conflate the quality of the PI with the difficulty of the predictive task at hand (see Section \ref{Sec:RelativeGain}). We set two desiderata:

{\bf Operating Point (OP) Variety.}
The importance of OP-comprehensive evaluation metrics is well understood, as demonstrated by techniques such as ROC curves \cite{Fawcett06}. 
In the context of PI, we define the term OP as a specific setting producing a certain value of mean coverage and bandwidth (a formal definition will be given in Section \ref{Sec:UCC}).

{\bf Interpretability Across Datasets}
When possible, metrics should capture the effectiveness of the technique being evaluated, rather than characterize the underlying dataset used in the evaluation.  

This paper proposes a methodology for evaluating prediction intervals that addresses both desiderata.  First, we introduce the Uncertainty Characteristics Curve (UCC), which leverages the well known concepts of operating characteristic curves to enable OP-comprehensive evaluation. Second, we introduce the notion of a {\em gain} over a null reference, which intuitively captures the quality of a prediction interval and allows for a meaningful comparison across different models as well as datasets.

\vspace{-.2cm}
\section{Method}

\vspace{-.15cm}
\subsection{Metrics}
\label{Sec:Metrics}
Suppose there are two components of a regression model: one generating target predictions, $\hat{Y}$, the other assigning uncertainty (lower and upper) bounds, $\hat{Y}^l, \hat{Y}^u$. 
Let $V=[Y, \hat{Y}, \hat{Y}^l, \hat{Y}^u]^T\in\mathbb{R}^4$ 
denote a multivariate random variable comprising the (observed) ground truth, $Y$, and the above-mentioned predictions. We also denote by $\hat{Z}^l, \hat{Z}^u$ the predicted lower an upper {\em bands}, s.t. $\hat{Y}^{l}=\hat{Y}-\hat{Z}^{l}$ and $\hat{Y}^{u}=\hat{Y}+\hat{Z}^{u}$. For simplicity we assume the regression task involves one-dimensional output.

\begin{figure}[tb]
  \centering
  \includegraphics[height=3.5cm]{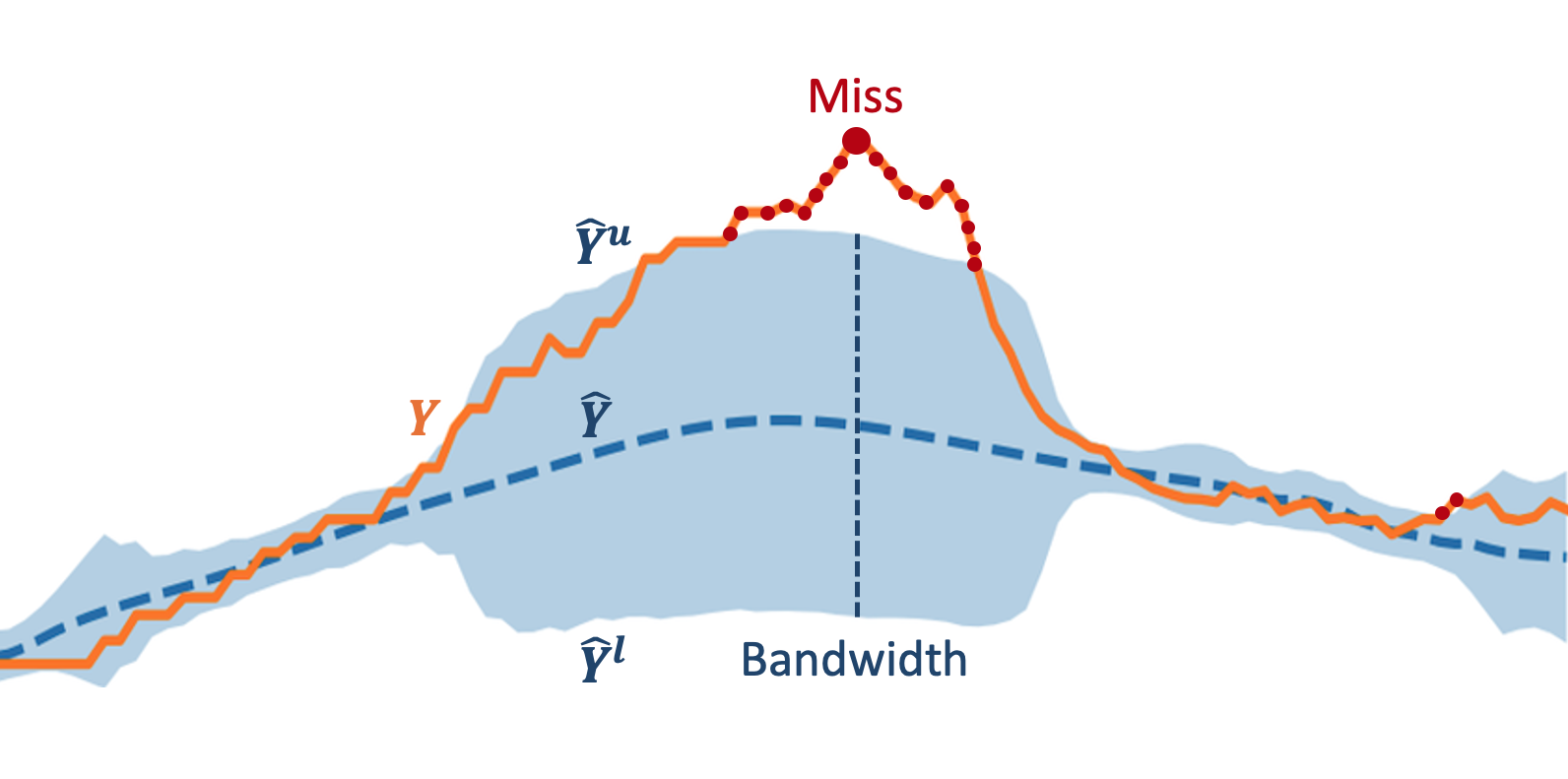}
  \captionof{figure}{Essential quantities on an illustrative (sequential) data example: $Y$ (observation), $\hat{Y}$ (regression prediction), $\hat{Y}^{l,u}$ (lower and upper bound), along with the two costs (a miss and a bandwdith) at a particular point of measurement. 
  The metrics average these costs over the data batch.
  }
  \label{Fig:TwoCosts}
\end{figure}

Fundamentally, given a batch of data, two costs arise in the assessment of prediction intervals:  (1) extent of observations falling outside the uncertainty bounds (miss rate), and (2) the average width of the bounds. These
two costs are in a trade-off relationship: wider intervals tend to reduce miss rates and vice versa.  
An illustrative example is shown in Figure \ref{Fig:TwoCosts} indicating the essential quantities and the two costs incurred at a particular point of the measurement. The total batch metrics are calculated as an average over the individual measurements.  
More formally, given a batch of $N$ measurements, $\mathbf{v}=\{v_1, ..., v_N\}$, where 
$v_i=(y_i, \hat{y}_i, \hat{y}^l_i, \hat{y}^u_i)$ being a realization of $V$, we define the metrics as follows:
\vspace{-.3cm}
\begin{equation}
    \mbox{Miss rate: }
    \qquad\hat{\rho}(\mathbf{v}) =
    \frac{1}{N}\sum_{i:y_{i}\notin[\hat{y}_{i}^l, \hat{y}_{i}^u]} 1 \label{Eq:Missrate}
\end{equation}
\vspace{-.3cm}
\begin{equation}
    \mbox{Bandwidth: }
    \qquad \hat{\beta}(\mathbf{v}) =
    \frac{1}{2N}\sum_{i=1}^{N} \hat{y}_{i}^u-\hat{y}_{i}^l \label{Eq:Bandwidth}
\end{equation}
We use the hat to indicate that the above are empirical averages. 

In case the variables $Y, \hat{Y}, \hat{Z}^l$, and $\hat{Z}^u$ are multi-dimensional, the above metrics may be calculated as averages over the individual dimensions.

{\subsection{The Uncertainty Characteristics Curve}
\label{Sec:UCC}}

The definition of the Uncertainty Characteristics Curve (UCC) hinges on a 
key scaling concept controlling the operating point (OP) which is described first.

\subsubsection{Scaling}\label{Sec:Scaling}
In general, the goal of setting an operating point (OP) is to transform the bounds such that a certain proportion of future observations, in expectation, falls within these bounds (aka calibration). 
Numerous techniques have been developed in the context 
of regression, e.g., \cite{KuleshovAccurateUncertainties, SongDistributionCalibration2019}. 
Employing the idea of Conformal Prediction \cite{VovkBook2005,Romano2019} we demonstrate that scaling 
plays a fundamental role in varying the OP. 
Conformal Prediction (CP) uses calibration data, $(X_1,Y_1),...,(X_n,Y_n)$ to produce a prediction interval $T(X)$ 
that is valid in the following sense: $1-\alpha\leq P(Y_{n+1}\in T(X_{n+1}))\leq 1-\alpha + \frac{1}{n+1}$, where 
$(X_{n+1},Y_{n+1})$ is a ``fresh''sample from same distribution as the calibration set, and $1-\alpha$ is the desired confidence level. A valid $T(X)$ will contain the new sample, $Y_{n+1}$ almost exactly with the desired probability. 
There are several CP variants achieving that. In general, a CP variant depends on a choice of a score function, $s(X,Y)$ that reflects the extent of a model's uncertainty.
Scores, $s_1, ..., s_n$ are calculated on the calibration set
and $\hat{q}$ as their $\frac{\ceil{(n+1)(1-\alpha)}}{n}$ quantile is determined. The PI then becomes 
$T(X)=\{y: s(X,y)\leq \hat{q}\}$ and the above CP probability guarantee holds. 
In regression tasks, it can be shown \cite{Romano2019} that by choosing $S(X,Y)=\frac{|Y-\hat{Y}|}{\hat{Z}}$ where $\hat{Z}$ is
an arbitrary function reflecting the uncertainty band at $X$, 
(e.g., an estimated standard deviation in case of a Gaussian), 
the prediction interval will be an interval $T(X)=[\hat{Y}-\hat{Z}\hat{q}, \hat{Y}+\hat{Z}\hat{q}]$.
In other words, the scaler $\hat{q}$ acts as a multiplicative correction factor applied to the uncertainty band, $\hat{Z}$, where larger values induce wider prediction intervals and thus smaller miss rates, and vice versa. The scaler $\hat{q}$ serves as an important element in controlling the overall bandwidth. 

\subsubsection{The Curve}
Given a data batch, $\mathbf{v}$, a specific scaling value ($\hat{q}$ above) induces a specific miss rate and bandwidth and thus characterizes a particular Operating Point (OP). A {\em set} of OPs can then be obtained by {\em varying} the scale applied to ${\hat{Z}}^l$ and ${\hat{Z}}^u$ over a range relevant to the data $\mathbf{v}$. 
With $k> 0$ denoting the scaling variable, we recast the dataset $\mathbf{v}$ as a function of $k$: 
\begin{eqnarray}
    \mathbf{v}(k) = \left\{v_i(k)\right\}_{1\leq i\leq N}=\nonumber\\
    =\left\{[y_i, \hat{y}_i, \hat{y}_i-k\hat{z}_i^l, \hat{y}_i+k\hat{z}_i^u]^T\right\}_{1\leq i\leq N}.
\end{eqnarray}
where $\hat{z}_i^l=\hat{y}_i - \hat{y}_i^l$ and $\hat{z}_i^u=\hat{y}_i^u - \hat{y}_i$ are the predicted bands for a sample $i$, 
scaled by the variable $k$. 
To further simplify the notation, we use a shorthand to write the bandwidth and the miss rate as functions of $k$
\begin{equation}
    \hat{\beta}(k):=\hat{\beta}(\mathbf{v}(k))\enspace\mbox{and}\enspace
    \hat{\rho}(k):=\hat{\rho}(\mathbf{v}(k)).
    \label{Eq:MetricShorthand}
\end{equation} 
E.g., $\hat{\rho}(k)$ gives the average miss rate of a batch after re-scaling the uncertainty bands, 
$\hat{z}_i^{l,u}$, using $k$. It can be 
readily observed that $\hat{\beta}(k)=c\cdot k$ with $c$ a constant depending on the original dataset. 

We now define the Uncertainty Characteristics Curve. 
\begin{definition}
    The Uncertainty Characteristics Curve (UCC) is a set of operating points $\left\{\left(\hat{\beta}(k), \hat{\rho}(k)\right)\right\}_{k\in S}$
    forming a bidimensional graph with the x-axis corresponding to the bandwidth and the y-axis  
    to the miss rate, and with $S$ a set of desirable scales. 
\end{definition}

The UCC graph shows the {\em trade-off} between the two cost metrics as a function of the scaling $k$.
As with the ROC \cite{Fawcett06}, a UCC can be parametric, however, in most practical cases is considered non-parametric with the cardinality of $S$ determined by the number of observations.

Given a dataset of size $N$, Algorithm \ref{Alg:UCC} shows an efficient computation of the UCC with a complexity of $O(N^2)$. For each sample, a critical scale is determined to adjust (widen or shrink) the interval bandwidth to just capture the ground truth with no excess.   
\begin{algorithm}[b]
  \caption{Algorithm to calculate the UCC}
  \label{Alg:UCC}
\begin{algorithmic}
    \State {\bfseries Input:}
  Ground truth, predictions, uncertainty estimates $\{y_i, \hat{y}_i, \hat{y}_i^l, \hat{y}_i^u\}_{1\leq i\leq N}$ 
    \State {\bfseries Output:} Set of UCC points $\{x_i, y_i\}_{1\leq i\leq N}$
    \For{$i\leftarrow 1$ {\bfseries to} $N$}
    \State $z_i\leftarrow y_i - \hat{y}_i$ \Comment{Observed error}
    \State 
    $k_i \leftarrow   
     \begin{cases}
      \frac{z_i}{\hat{y}_i^u-\hat{y}}& \text{for}\,\,z_i\geq 0\\
      -\frac{z_i}{\hat{y}-\hat{y}_i^l}& \text{otherwise}\\
     \end{cases}
    $ \Comment{Critical scale} 
    \State $x_i \leftarrow \hat{\beta}(k_i)$  \Comment{$\hat{\beta}, \hat{\rho}$ defined in Eq. (\ref{Eq:MetricShorthand})}
    \State $y_i \leftarrow \hat{\rho}(k_i)$
    \EndFor
\end{algorithmic}
\end{algorithm}

\begin{figure}[tb]
  \centering
  \includegraphics[height=3.5cm, width=5.5cm]{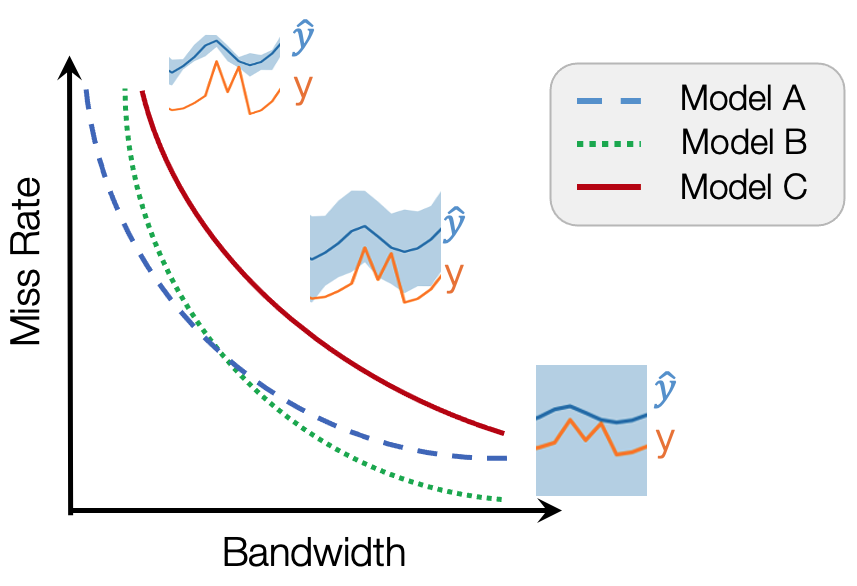}
  \captionof{figure}{An illustrative example of multiple UCCs obtained for different models. Highlighted are three different operating regimes: High (Top), Medium (Middle), and Low (Bottom) Miss Rate.  }
  \label{Fig:UCCExample1}
\end{figure}
An illustrative example of a UCC graph is given in Figure \ref{Fig:UCCExample1} showing three curves corresponding to three different, hypothetical models generating uncertainty bounds $\hat{z}$ around the same target predictions $\hat{y}$. Illustrative icons characterizing low, middle, and high miss rate regimes are also shown. Each UCC curve reflects the operating characteristics of its model by showing a trade-off between the two costs. In this example, the curve for model C dominates A and B and is therefore inferior as it implies higher bandwidth is needed to achieve any given miss rate. In contrast, the model A appears superior to B in a low bandwidth range, while B outperforms A in a low miss rate area. Note that each curve eventually intercepts both axes reaching a zero value. 

Similar to the ROC in detection tasks, a UCC can reveal a model's suitability for certain operating regimes, and offers a way to compare several models in an application-agnostic way.

\subsubsection{Area Under the UCC (AUUCC)}

It is desirable to define a summary metric capturing the overall quality of a model generating prediction intervals. Given that the UCC coordinates correspond to costs, a sensible choice is the area under the curve (or AUUCC). Models generating bounds that incur lower cost across the entire 
operating range will produce a lower AUUCC. Thus, in absence of a pre-determined operating point, the AUUCC measure is a useful OP-agnostic summary. Alternatively, if a certain range of a cost (e.g., the miss rate) is anticipated, a {\em partial} AUUCC focusing on that range can be determined, similar to the notion of partial ROC AUC \cite{PartialAUC2013}.

Unlike with the ROC AUC analysis, the range of the AUUCC depends on the PI range and is therefore data dependent. This underscores the need for normalization as discussed next.

\subsubsection{Null Reference}\label{Sec:RelativeGain}

While we want to assess the quality of the {\em uncertainty}, most 
standard metrics (e.g., likelihood, bandwidth, etc.) confound  
uncertainty bands with actual target predictions. To illustrate, suppose there is a model predicting a regression target, $\hat{y}$, as well as a gaussian uncertainty (variance) $\hat{\sigma}^2$. The loss with respect to model parameters $\theta$ (the negative log likelihood) is a function of both the predictions, $\hat{y}_i$ and the uncertainty, $\hat{\sigma}_i$:
$-log P(\theta) = \frac{1}{2} \sum_i^N \frac{(y_i-\hat{y}_i)^2}{\hat{\sigma}_i^2} + \log \hat{\sigma}_i^2 + const$.
This metric, often used to make judgements in uncertainty 
quantification \cite{ovadia2019trustinuncertainty, Kendall2017_whatuncertainty, Oh2020_crowdcounting}, entangles both qualities (target regression accuracy and prediction interval quality). 
As a consequence, in their raw form, metrics like these (incl. the bandwidth, etc) are difficult to  
compare across different regressors. Therefore we 
look for a {\em relative gain} of our metrics over a simple, intuitive reference. A suitable choice of such a reference 
are {\em constant bands}, i.e. $\forall i:\hat{z}_i^l=const, \hat{z}_i^u=const$. Given target predictions, 
$\hat{y_i}$, such a non-informative (null) reference is always possible to generate and often represents a plausible, effective choice.

\subsubsection{AUUCC Gain}

Let $A_M$ represent the AUUCC of a model and let $A_{Const}$ refer to the constant-band counterpart.
We define the AUUCC Gain as follows:
\begin{equation}
    G_M = \frac{A_{Const}-A_M}{A_{Const}}\cdot 100\%\label{Eq:AUUCCGain}.
\end{equation}
A positive value summarizes the overall quality in an OP-independent manner. 
Negative gains are an indication of a model issue (misspecification, over-training, etc.). 
The partial-AUUCC gain is calculated similarly via Eq. (\ref{Eq:AUUCCGain}) using partial AUUCC values. 

\begin{figure}[tbp!]
       \centering
       \includegraphics[height=4.5cm]{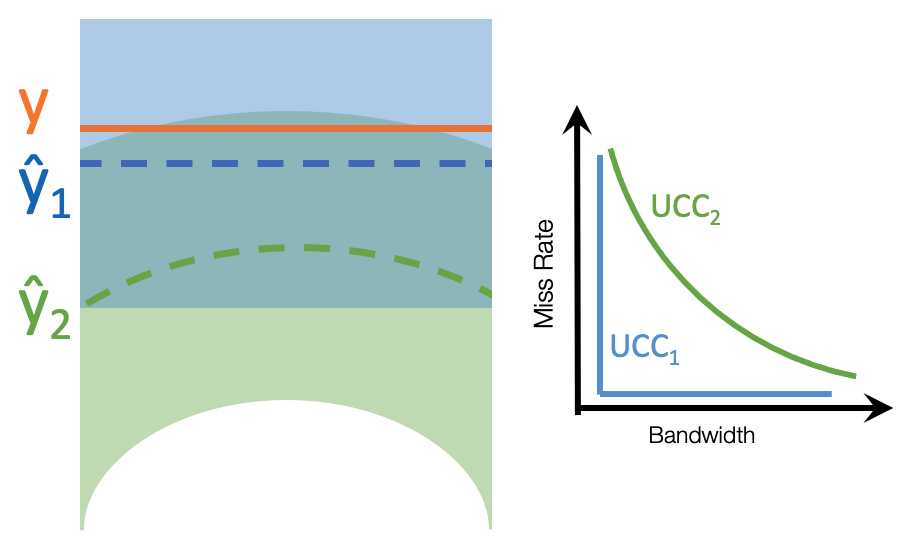}
       \caption{Illustrative example of how identical PI--constant bands in both cases--lead to different UCCs if they relate to different target predictions, $\hat{y}$}
       \label{Fig:yhatmatters}
\end{figure}

The benefit of the gain metric can be illustrated on a simple example: two prediction intervals are shown in Figure \ref{Fig:yhatmatters} (shaded in blue and green). Both are constant bands, i.e., their expected quality should be same. However, 
because they relate each to different target predictions, $\hat{y_1}$ and $\hat{y_2}$, their 
behavior with respect to the observation, $y$, is different. While the interval around $\hat{y_1}$ captures the observation fully at a certain critical scale leading to a perfect curve (UCC1), the interval around $\hat{y_2}$ incurs a positive
miss rate even at the same bandwidth. As the bandwidth varies, the characteristics of the curved band 
intersecting the observation will be non-trivial leading to the (inferior) curve UCC2 as illustrated. 
This example highlights the need for normalization  
(see Eq. (\ref{Eq:AUUCCGain})) which would result in a gain of 0\% for both cases thus making the assessment of the intervals equivalent in terms of their quality.

A probabilistic interpretation of the AUUCC exists and is included in the Appendix.

\subsubsection{Significance Testing}
 Standard tools of significance testing are applicable to the UCC. 
 If a pairwise comparison between two models in terms of the AUUCC is desired, the non-parametric 
 paired permutation test \cite{dwass1957_permutationtest} is applicable.

\section{Related Work}

Uncertainty quantification in machine learning is a long-standing field of active research. 
Sources
of uncertainty are generally categorized as epistemic or aleatoric \cite{Kiureghian09, Kendall2017_whatuncertainty}. In classification tasks, uncertainty is expressed as a measure of confidence  
accompanying a result \cite{gal2015theoretically, Guo2017, Lakshminarayanan2017, ovadia2019trustinuncertainty}.
Combined with an optional calibration step, e.g. \cite{Zadrozny2002}, a quality assessment of such estimates relies on summary metrics, such as the Brier score \cite{Brier1951, Bradley2019, Lakshminarayanan2017, Guo2017, Zadrozny01obtainingcalibrated}, Expected Calibration Error \cite{Naeini15, Guo2017}, ROC-like metrics, and  Accuracy-vs-Confidence curves \cite{Chen2019_linearprobes, Lakshminarayanan2017}.
Uncertainty in {\em regression} tasks involves estimating prediction intervals 
(e.g., \cite{KoenkerQuantileRegression78, Papoulis1989}) as well as 
in state-of-the-art machine learning \cite{Nix1994_variancemodel, gal2015theoretically, Kendall2017_whatuncertainty}. 
However, the methodology of comparing their quality is relatively scarce, ranging  
from reliance on calibration and sharpness \cite{KuleshovAccurateUncertainties, Gneiting2007}, coverage  \cite{Oh2020_crowdcounting}, to using summary likelihood measures \cite{Lakshminarayanan2017}.

The Uncertainty Characteristic Curve (UCC) broadens the evaluation aspect drawing an analogy to the well-known ROC \cite{Fawcett06}.
The trade-off between two costs has been studied and applied previously \cite{Gneiting2007, Dunsmore1968, Shen2018_mentionsbandwidth, Tagasovska2019}. However, most reports rely on a specific OP during the assessment stage. 
In the context of regression, Bi et al. \cite{Bi_REC_2003} developed an assessment tool termed Regression Error Characteristic (REC) curve utilizing a constant tolerance band around a regression target. The REC allows for a comprehensive assessment of regressors. The UCC is conceived in a similar vein. Besides the different application and metrics used, 
the UCC fundamentally differs from the REC in not relying on varying a constant tolerance band but generalizing 
to an arbitrary-shape tolerance band via scaling. In this sense, the UCC is a generalization of the REC.
Finally, our work should be contrasted to calibration curves (also known as 
reliability diagrams) used for assessment primarily in prognostic aspects of classification tasks \cite{ReliabilityDiagrams2005}, and, more recently, in regression tasks \cite{tran2020methods}.
A calibration curve captures the amount of over- and under-confidence in the PI with respect to observed 
quantiles. While these curves vary the calibration setting 
there are two essential differences: (1) both axes are quantiles (i.e., there is no cost trade-off relationship), 
and (2) the actual quality ("accuracy") of the PI is not captured. Poor
PI can obtain a perfect calibration curve and vice versa. 

\section{Experiments}

In this section we present two case studies highlighting interesting scenarios where UCCs offer crucial insights.

\vspace{-.3cm}

\subsubsection{Synthetic Data Example}


\begin{figure*} [htb]
      \centering
      \includegraphics[width=14cm, height=4cm]{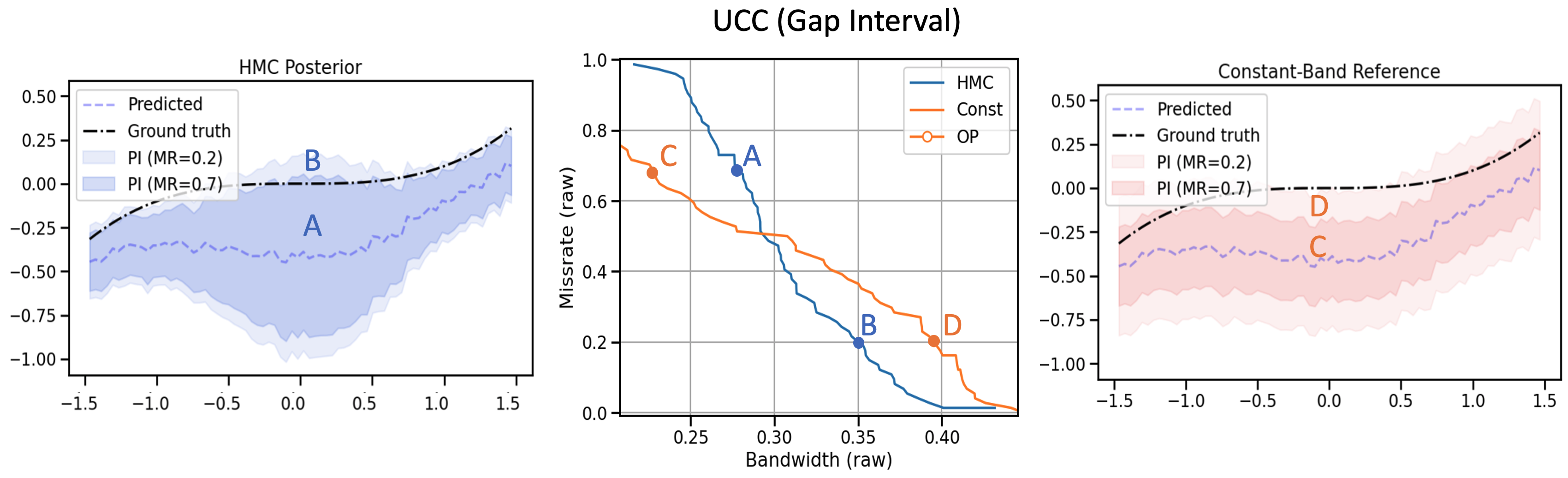}
      \caption{Synthetic-data example with BNN predictions (left), the UCC cross-over (middle), and the constant-band reference (right). MR stands for Miss Rate.}
      \label{Fig:ZoomGap}
\end{figure*}
Which of the prediction intervals shown in Figure \ref{Fig:ZoomGap} should be considered better?
This example illustrates the complexities in prediction interval assessment. 
A Bayesian Neural Network (BNN) is trained on 
a one-dimensional synthetic regression task involving  
a data gap in the region $[-1,1]$, similar to \cite{yao2019quality}. 
The Hamiltoninan Monte Carlo (HMC) inference method---largely considered the gold standard for BNNs \cite{neal2011mcmc,Foong2020}---is used. The outcome in terms of prediction, ground truth (observation), 
and prediction intervals (PI) are shown in Figure 
\ref{Fig:ZoomGap} (left-most plot). The PIs are shown for two miss rate settings (adjusted via appropriate scaling): 0.2 (labeled as B) and 0.7 (labeled A). 
A significant widening in the uncertainty occurs within the region $[-1,1]$ caused by the gap in training, which is consistent with our intuition. 
The corresponding constant-band reference is shown in the right-most plot of Figure \ref{Fig:ZoomGap}, for 
the same 0.2 (labeled D) and 0.7 (labeled C) miss rate settings. 
Which of these prediction intervals (HMC or Constant) should be considered superior?
It turns out this question cannot be answered without a careful assessment.
The middle plot in Figure \ref{Fig:ZoomGap} shows the UCC chart comprising two curves: one for the HMC PIs (blue)
and one for the constant-band reference (orange). 
The two distinct OPs are mapped on each curve. OPs A and C (corresponding to the left- and right-most plots) 
lie in at the high miss rate 
of 0.7, and OPs B and D lie low at 0.2. 
Comparing the two PIs at the miss rate of 0.7, the UCC indicates that the constant band incurs lower bandwidth cost 
thus outperforms the HMC. However, comparing the two at the OP B and D, the reverse is true. In other words, each model 
can outperform the other, depending on the operating region. 
This "cross-over" observation can be explained by a closer look at the two data plots: while in the setting A the HMC 
misses 70\% of the ground truth and has larger bandwidth than the reference, in the setting B (20\% miss rate) the HMC
benefits from the interval widening in the center allowing it to spend its bandwidth more efficiently.  
This example underscores the insight in the UCC:
Evaluating the above PIs at any fixed operating point would tell an incomplete story, the conclusion of which depends on a particular operating point chosen.

The overall AUUCC gain of the HMC PI over its constant reference ($G_M$ as per Eq. (\ref{Eq:AUUCCGain})) is 6.1\%. 
This is relatively low and reflects the mixed outcome seen in the UCC chart. The gain changes 
dramatically when calculating a {\em partial} AUUCC gain focusing on the miss rate range between 0 and 0.5: 
the partial gain of the HMC grows to 72.7\% clearly indicating superiority of the HMC in this lower miss rate range.

\subsubsection{Traffic Volume Prediction}
\label{Sec:RealWorldDatasets}
\vspace{-.3cm}
To exercise the UCC on a real-world task we selected the Metro Interstate Traffic Volume Dataset\footnote{https://archive.ics.uci.edu/ml/machine-learning-databases/00492/}, 
which is a sequential regression task with 48204 traffic volume observations (as the regression variable)
along with weather conditions (as the covariates). 
The predictor is an LSTM-based sequence-to-sequence architecture developed for this task in \cite{SRT2021Arxiv} 
generating both the target predictions as well as the uncertainty intervals using a joint meta-modeling approach. 

\begin{figure} [htb]
       \centering
       \includegraphics[height=4.cm]{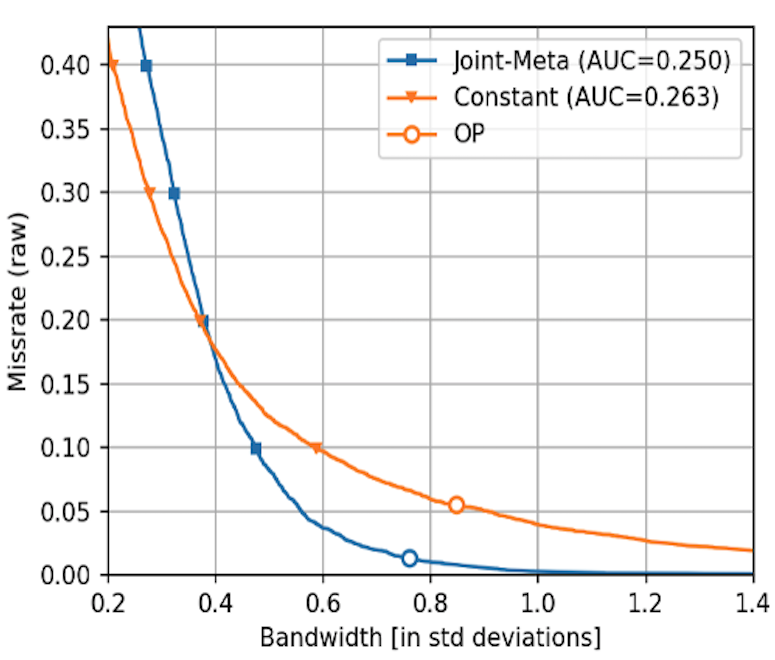}
       \caption{UCC on the Traffic dataset using an LSTM-based Joint Meta-Model predictor \cite{SRT2021Arxiv}}
       \label{Fig:TrafficUCC}
\end{figure}
Figure \ref{Fig:TrafficUCC} shows the UCC of the model's PI ("Joint-Meta") along with its constant-band reference ("Constant"). 
The model outperforms its constant baseline in a low miss rate range ($< 0.2$) but falls short 
anywhere above that value. The overall AUUCC gain is only 4.9\%. From our broader experimentation 
it appears that such cross-over phenomena occur in many real-world datasets with cross-over 
locations varying widely. These findings underscore the need for a visualization and a comprehensive assessment (the UCC)
otherwise phenomena such as these may be easily missed when evaluated at an ad-hoc operating point.

It may be argued that certain UCC ranges may not be of practical interest, for example, miss rates of 50\% or more may be considered too high. However, some applications, such as anomaly detection, utilize predicted uncertainty bands in conjunction with observations to detect anomalous events (observations falling outside of the band). In some cases, it may be sensible to operate in a relatively high miss rate mode as missing an anomalous event may be costly. 
We believe that, as an OP-agnostic tool, the UCC should include the full range in absence of  a-priori knowledge, however, the UCC analysis, in particular the AUUCC, can be adapted to a partial area as discussed above. 

\vspace{-.2cm}
\paragraph*{UCC Implementation}

The UCC was implemented in Python 3 and is publicly available as part of the Uncertainty Quantification 360 Toolkit\footnote{https://github.com/IBM/UQ360, http://uq360.mybluemix.net} \cite{UQ360Tutorial2022}.

\vspace{-.2cm}
\section {Conclusions}
\vspace{-.2cm}
In this work we introduced the Uncertainty Characteristics Curve (UCC) in conjunction with 
a gain metric relative to constant-band references, and demonstrated its benefit in diagnostics of prediction intervals. The UCC is formed by varying a scaling-based operating point manipulation applied to the prediction intervals,
thus characterizing their quality in an operating point agnostic manner. 
In two case studies, the UCC 
was shown to provide important insights in terms of both the AUUCC gain metrics and the operating characteristics along the operating range. 
With the corresponding code available, we believe the UCC will become a valuable new addition in the diagnostic toolbox for ML researchers and practitioners alike.  
\raggedbottom
\bibliographystyle{abbrv}
\bibliography{main}

\begin{thebibliography}{10}

\bibitem{Arnold2019_factsheets}
M.~{Arnold}, R.~K.~E. {Bellamy}, M.~{Hind}, S.~{Houde}, S.~{Mehta},
  A.~{Mojsilovi\'c}, R.~{Nair}, K.~N. {Ramamurthy}, A.~{Olteanu},
  D.~{Piorkowski}, D.~{Reimer}, J.~{Richards}, J.~{Tsay}, and K.~R. {Varshney}.
\newblock Factsheets: Increasing trust in ai services through supplier's
  declarations of conformity.
\newblock {\em IBM Journal of Research and Development}, 63(4/5):6:1--6:13,
  July 2019.

\bibitem{Begoli2019}
E.~Begoli, T.~Bhattacharya, and D.~Kusnezov.
\newblock The need for uncertainty quantification in machine-assisted medical
  decision making.
\newblock {\em Nature Mach Intell}, 1:20--23, 2019.

\bibitem{Bi_REC_2003}
J.~Bi and K.~P. Bennett.
\newblock Regression error characteristic curves.
\newblock ICML’03, page 43–50. AAAI Press, 2003.

\bibitem{Bradley2019}
A.~A. Bradley, J.~Demargne, and K.~J. Franz.
\newblock {\em Attributes of Forecast Quality}, pages 849--892.
\newblock Springer Berlin Heidelberg, Berlin, Heidelberg, 2019.

\bibitem{Brier1951}
G.~W. Brier and R.~A. Allen.
\newblock {\em Verification of Weather Forecasts}, pages 841--848.
\newblock American Meteorological Society, Boston, MA, 1951.

\bibitem{CasellaBook}
G.~Casella and R.~Berger.
\newblock {\em Statistical Inference}.
\newblock {Duxbury Resource Center}, June 2001.

\bibitem{Chen2019_linearprobes}
T.~Chen, J.~Navr{\'{a}}til, V.~Iyengar, and K.~Shanmugam.
\newblock Confidence scoring using whitebox meta-models with linear classifier
  probes.
\newblock In {\em {AISTATS} 2019, 16-18 April 2019, Naha, Okinawa, Japan},
  pages 1467--1475, 2019.

\bibitem{Kiureghian09}
A.~{Der Kiureghian} and O.~Ditlevsen.
\newblock Aleatoric or epistemic? does it matter?
\newblock {\em Structural Safety}, 31(2):105--112, 2009.

\bibitem{Dunsmore1968}
I.~R. Dunsmore.
\newblock A bayesian approach to calibration.
\newblock {\em Journal of the Royal Statistical Society: Series B
  (Methodological)}, 30(2):396--405, 1968.

\bibitem{dwass1957_permutationtest}
M.~Dwass.
\newblock Modified randomization tests for nonparametric hypotheses.
\newblock {\em Ann. Math. Statist.}, 28(1):181--187, 03 1957.

\bibitem{Fawcett06}
T.~Fawcett.
\newblock An introduction to roc analysis.
\newblock {\em Pattern Recognition Letters}, 27(8):861 -- 874, 2006.
\newblock ROC Analysis in Pattern Recognition.

\bibitem{Foong2020}
A.~Foong, D.~Burt, Y.~Li, and R.~Turner.
\newblock On the expressiveness of approximate inference in bayesian neural
  networks.
\newblock In {\em Advances in Neural Information Processing Systems},
  volume~33, pages 15897--15908, 2020.

\bibitem{gal2015theoretically}
Y.~Gal and Z.~Ghahramani.
\newblock A theoretically grounded application of dropout in recurrent neural
  networks.
\newblock In {\em Advances in Neural Information Processing Systems 29}, pages
  1019--1027. Curran Associates, Inc., 2016.

\bibitem{UQ360Tutorial2022}
S.~Ghosh, Q.~V. Liao, K.~Ramamurthy, J.~Navratil, P.~Sattigeri, K.~Varshney,
  and Y.~Zhang.
\newblock Uncertainty quantification 360: A hands-on tutorial.
\newblock CODS-COMAD 2022, 2022.

\bibitem{Gneiting2007}
T.~Gneiting, F.~Balabdaoui, and A.~E. Raftery.
\newblock Probabilistic forecasts, calibration and sharpness.
\newblock {\em Journal of the Royal Statistical Society: Series B (Statistical
  Methodology)}, 69(2):243--268, 2007.

\bibitem{Guo2017}
C.~Guo, G.~Pleiss, Y.~Sun, and K.~Q. Weinberger.
\newblock On calibration of modern neural networks.
\newblock ICML’17, page 1321–1330. JMLR.org, 2017.

\bibitem{Jiang2018_trust}
H.~Jiang, B.~Kim, M.~Guan, and M.~Gupta.
\newblock To trust or not to trust a classifier.
\newblock In {\em Advances in Neural Information Processing Systems 31}, pages
  5541--5552. Curran Associates, Inc., 2018.

\bibitem{Kendall2017_whatuncertainty}
A.~Kendall and Y.~Gal.
\newblock What uncertainties do we need in bayesian deep learning for computer
  vision?
\newblock In {\em Advances in Neural Information Processing Systems 30}, pages
  5574--5584. Curran Associates, Inc., 2017.

\bibitem{KoenkerQuantileRegression78}
R.~W. Koenker and G.~Bassett.
\newblock Regression quantiles.
\newblock {\em Econometrica}, 46(1):33--50, 1978.

\bibitem{KuleshovAccurateUncertainties}
V.~Kuleshov, N.~Fenner, and S.~Ermon.
\newblock Accurate uncertainties for deep learning using calibrated regression.
\newblock In {\em Proceedings of the 35th International Conference on Machine
  Learning}, volume~80 of {\em Proceedings of Machine Learning Research}, pages
  2796--2804, Stockholmsmässan, Stockholm Sweden, 10--15 Jul 2018. PMLR.

\bibitem{Lakshminarayanan2017}
B.~Lakshminarayanan, A.~Pritzel, and C.~Blundell.
\newblock Simple and scalable predictive uncertainty estimation using deep
  ensembles.
\newblock In {\em Advances in Neural Information Processing Systems 30}, pages
  6402--6413. Curran Associates, Inc., 2017.

\bibitem{Naeini15}
M.~P. Naeini, G.~F. Cooper, and M.~Hauskrecht.
\newblock Obtaining well calibrated probabilities using bayesian binning.
\newblock In {\em Proceedings of the Twenty-Ninth AAAI Conference on Artificial
  Intelligence}, AAAI’15, page 2901–2907. AAAI Press, 2015.

\bibitem{PartialAUC2013}
H.~Narasimhan and S.~Agarwal.
\newblock A structural {SVM} based approach for optimizing partial auc.
\newblock volume~28 of {\em Proceedings of Machine Learning Research}, pages
  516--524, Atlanta, Georgia, USA, 17--19 Jun 2013. PMLR.

\bibitem{neal2011mcmc}
R.~M. Neal et~al.
\newblock Mcmc using hamiltonian dynamics.
\newblock {\em Handbook of markov chain monte carlo}, 2(11):2, 2011.

\bibitem{ReliabilityDiagrams2005}
A.~Niculescu-Mizil and R.~Caruana.
\newblock Predicting good probabilities with supervised learning.
\newblock In {\em Proceedings of the 22nd International Conference on Machine
  Learning}, ICML '05, page 625–632, New York, NY, USA, 2005. Association for
  Computing Machinery.

\bibitem{Nix1994_variancemodel}
D.~A. {Nix} and A.~S. {Weigend}.
\newblock Estimating the mean and variance of the target probability
  distribution.
\newblock In {\em Proceedings of 1994 IEEE International Conference on Neural
  Networks (ICNN'94)}, volume~1, pages 55--60 vol.1, June 1994.

\bibitem{Oh2020_crowdcounting}
M.~Oh, P.~A. Olsen, and K.~N. Ramamurthy.
\newblock Crowd counting with decomposed uncertainty.
\newblock In {\em AAAI Conference on Artificial Intelligence}, 2020.

\bibitem{Papoulis1989}
A.~Papoulis and H.~Saunders.
\newblock {Probability, Random Variables and Stochastic Processes (2nd
  Edition)}.
\newblock {\em Journal of Vibration, Acoustics, Stress, and Reliability in
  Design}, 111(1):123--125, 01 1989.

\bibitem{Romano2019}
Y.~Romano, E.~Patterson, and E.~Candes.
\newblock Conformalized quantile regression.
\newblock In {\em Advances in Neural Information Processing Systems},
  volume~32. Curran Associates, Inc., 2019.

\bibitem{Shen2018_mentionsbandwidth}
Y.~Shen, X.~Wang, and J.~Chen.
\newblock Wind power forecasting using multi-objective evolutionary algorithms
  for wavelet neural network-optimized prediction intervals.
\newblock {\em Applied Sciences}, 8(2):185, Jan 2018.

\bibitem{ovadia2019trustinuncertainty}
J.~Snoek, Y.~Ovadia, E.~Fertig, B.~Lakshminarayanan, S.~Nowozin, D.~Sculley,
  J.~Dillon, J.~Ren, and Z.~Nado.
\newblock Can you trust your model's uncertainty? evaluating predictive
  uncertainty under dataset shift.
\newblock In {\em Advances in Neural Information Processing Systems}, pages
  13969--13980, 2019.

\bibitem{SongDistributionCalibration2019}
H.~Song, T.~Diethe, M.~Kull, and P.~Flach.
\newblock Distribution calibration for regression.
\newblock In {\em International Conference on Machine Learning, 9-15 June 2019,
  Long Beach, California, USA}, Proceedings of Machine Learning Research, pages
  5897--5906. Proceedings of Machine Learning Research, May 2019.

\bibitem{Tagasovska2019}
N.~Tagasovska and D.~Lopez-Paz.
\newblock Single-model uncertainties for deep learning.
\newblock In H.~Wallach, H.~Larochelle, A.~Beygelzimer, F.~d'~Alch\'{e}-Buc,
  E.~Fox, and R.~Garnett, editors, {\em Advances in Neural Information
  Processing Systems 32}, pages 6417--6428. Curran Associates, Inc., 2019.

\bibitem{tran2020methods}
K.~Tran, W.~Neiswanger, J.~Yoon, Q.~Zhang, E.~Xing, and Z.~W. Ulissi.
\newblock Methods for comparing uncertainty quantifications for material
  property predictions.
\newblock {\em Machine Learning: Science and Technology}, 1(2):025006, 2020.

\bibitem{VovkBook2005}
V.~Vovk, A.~Gammerman, and G.~Shafer.
\newblock {\em Algorithmic Learning in a Random World}.
\newblock Springer-Verlag, Berlin, Heidelberg, 2005.

\bibitem{SRT2021Arxiv}
XAuthors.
\newblock Anonymized.
\newblock In {\em Anonymized}, 2020.

\bibitem{yao2019quality}
J.~Yao, W.~Pan, S.~Ghosh, and F.~Doshi-Velez.
\newblock Quality of uncertainty quantification for bayesian neural network
  inference.
\newblock {\em ICML workshop on uncertainty in deep learning}, 2019.

\bibitem{Zadrozny01obtainingcalibrated}
B.~Zadrozny and C.~Elkan.
\newblock Obtaining calibrated probability estimates from decision trees and
  naive bayesian classifiers.
\newblock In {\em In Proceedings of the Eighteenth International Conference on
  Machine Learning}, pages 609--616. Morgan Kaufmann, 2001.

\bibitem{Zadrozny2002}
B.~Zadrozny and C.~Elkan.
\newblock Transforming classifier scores into accurate multiclass probability
  estimates.
\newblock In {\em Proceedings of the Eighth ACM SIGKDD International Conference
  on Knowledge Discovery and Data Mining}, KDD '02, pages 694--699, New York,
  NY, USA, 2002. ACM.

\end{thebibliography}
\pagebreak
\clearpage

\section{Appendix}

\subsection{Additional Metrics}\label{App:Sec:AddMetrics}
\begin{figure}[htbp!]
       \centering
       \includegraphics[height=2.5cm]{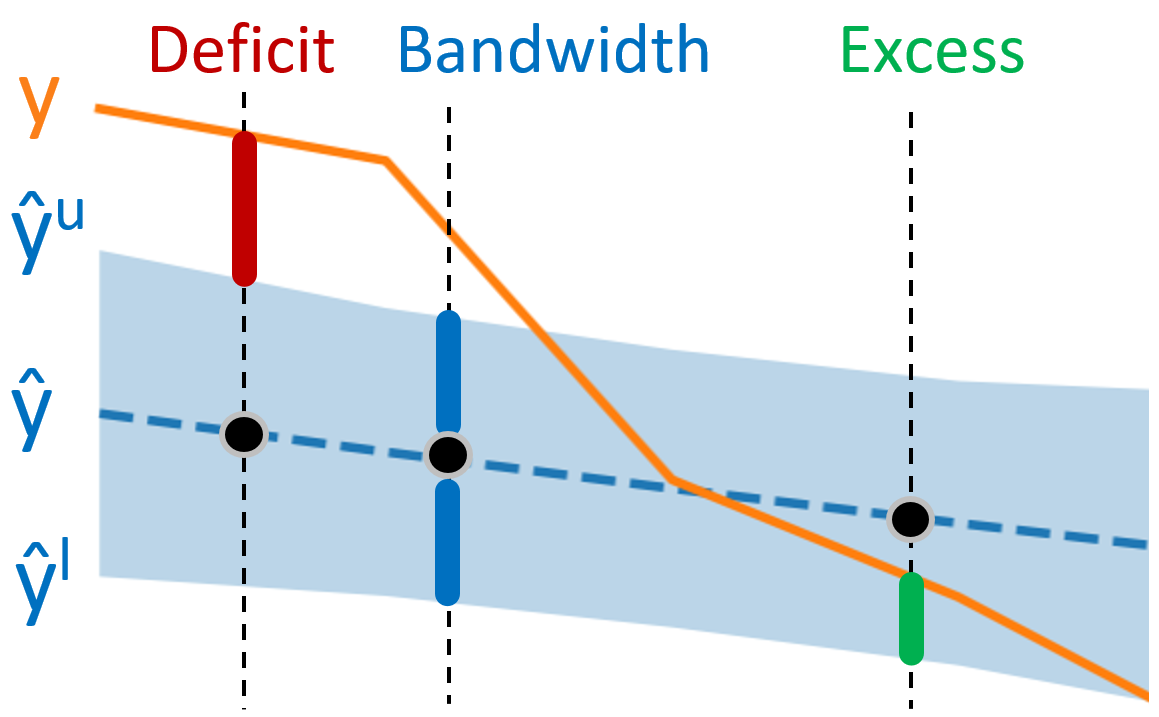}
       \caption{Bandwidth, excess, and deficit costs.}
       \label{Fig:Metrics}
\end{figure}

In addition to bandwidth and miss rate, defined in Section \ref{Sec:Metrics}, 
we propose two additional, related metrics as follows 
    \begin{flalign} 
        \mbox{Excess: }\nonumber\\
        \qquad\xi(V) &= \E\left[\mathbf{1}_{Y\in[\hat{Y}^l, \hat{Y}^u]}\cdot
            \min\left\{Y-\hat{Y}^l, \hat{Y}^u-Y\right\}\right]&\nonumber\\
        \qquad \hat{\xi}(\mathbf{v}) &=
        \frac{1}{N}\sum_{i:y_{i}\in[\hat{y}_{i}^l, \hat{y}_{i}^u]}
        \min\left\{y_{i}-\hat{y}_{i}^l, \hat{y}_{i}^u - y_{i}\right\}\label{Eq:Excess}&
    \end{flalign}
    \begin{flalign} 
        \mbox{Deficit: }\nonumber\\
        \qquad\delta(V)&=\E_{p_V}\left[\mathbf{1}_{Y\notin[\hat{Y}^l, \hat{Y}^u]}\cdot
            \min\left\{\left|Y-\hat{Y}^l\right|, \left|Y-\hat{Y}^u\right|\right\}\right]& \nonumber\\
        \qquad \hat{\delta}(\mathbf{v}) &=
        \frac{1}{N}\sum_{i:y_{i}\notin[\hat{y}_{i}^l, \hat{y}_{i}^u]}
        \min\left\{|y_{i}-\hat{y}_{i}^l|, |y_{i}  -\hat{y}_{i}^u|\right\}\label{Eq:Deficit}&
    \end{flalign}
Figure \ref{Fig:Metrics} illustrates all four metrics. The relative proportion of observations lying outside the bounds (i.e., the miss rate) ignores the {\em extent} of the bounds' shortfall. The proposed Deficit, Eq. (\ref{Eq:Deficit}), captures this aspect. The type 2 cost is captured by the Bandwidth, Eq. (\ref{Eq:Bandwidth}). However, its range is indirectly compounded by the underlying variation in ${\hat{Y}}$ and ${Y}$. Therefore we propose the Excess measure, Eq. (\ref{Eq:Excess}), which also reflects the Type 2 cost, but just the portion above the minimum bandwidth necessary to include the observation.
We will be using Excess and Deficit in reporting additional results in this document and also 
present an additional theoretical result for a UCC on Excess-Deficit coordinates.

\subsection{Proofs And Additional Results}
\label{App:Proofs}

A probabilistic interpretation of the area under the ROC is well known \cite{Fawcett06}. 
Bi et al. \cite{Bi_REC_2003} also established a connection between the area under the Regression Error Characteristics (REC) curve and the expected error. 
In a similar vein, we derive a probabilistic interpretation for the AUUCC.
\begin{definition}(Critical Scale).\label{Def:CritScale}
    Given an observation $v_i=[y_i, \hat{y}_i, \hat{y}_i^l, \hat{y}_i^u]^T$, 
    a critical scale is a factor $k_i$ calculated according to 
    \vspace{-.5cm}
    \begin{equation}
        k_i := 
     \begin{cases}
      \frac{z_i}{\hat{z}_i^u}& \text{for}\,\,z_i\geq 0\\
      -\frac{z_i}{\hat{z}_i^l}& \text{otherwise}\\
     \end{cases}\label{Eq:ScaleDef}
    \vspace{-.2cm}
    \end{equation}
    where $z_i=y_i - \hat{y}_i$, $\hat{z}_i^l=\hat{y}-\hat{y}_i^l$, and $\hat{z}_i^u=\hat{y}_i^u-\hat{y}$.
\end{definition}
The critical value $k_i$ scales the active (lower or upper) error band, $\hat{z}_i^{l,u}$, so that it captures the observation $y_i$ with no excess. Note that this notion is also utilized in 
the Algorithm \ref{Alg:UCC}.

Let $p_V$ denote the probability density of $V$ and $\mathbf{v}=\{v_1, ..., v_N\}$ a set of $N$ samples from $p_V$, where $v_i=[y_i, \hat{y}_i, \hat{y}_i^l, \hat{y}_i^u]^T$. 

\begin{proposition}\label{Prop:missrate}
    Let $B$ denote a bandwidth random variable generated by the following procedure: (i) Randomly select 
    an observation, $v=[y, \hat{y}, \hat{y}^l, \hat{y}^u]^T$ according to $p_V$, (ii) determine its critical scale, $k$, via Eq. (\ref{Eq:ScaleDef}), and (iii) obtain the average bandwidth value via
    Eq. (\ref{Eq:MetricShorthand}) using the exact metric $b={\beta}(k)$. Let $p_B$ denote the probability density of $B$.
    The area under the UCC, calculated from a finite sample $\{v_1,..., v_N\}$, is an estimator of the expected value $\left<B\right>_{p_B}$. 
\end{proposition}

The Proposition \ref{Prop:missrate} states that the AUUCC estimates the expected bandwidth over the set of data-induced operating points. Consequently, for a given sample of predictions, $\{v_1,..., v_N\}$, the sample average of the corresponding bandwidth values, $\{b_1,..., b_N\}$, determined in Algorithm \ref{Alg:UCC} approximates the AUUCC. This connection is analogous to one pointed out by Bi et al \cite{Bi_REC_2003} for the REC curve. 

To prove Proposition \ref{Prop:missrate} 
we use the Definition \ref{Def:CritScale} and the Lemma \ref{Lemma:1} below:

\begin{lemma}
    \label{Lemma:1}
    Choose any $v_i\in\mathbf{v}$, with $\mathbf{v}$ a sample set as defined in Section \ref{Sec:Metrics}.
    Let $k_i$ be the critical scale for $v_i$ and $K$ the scale random variable. The following holds
    \begin{equation}
        P(Y\notin[\hat{Y}-k_i\hat{Z}^l, \hat{Y}+k_i\hat{Z}^u])\equiv 1-P(K\leq k_i)\nonumber.
    \end{equation}
\end{lemma}
\begin{proof} Let $\{k_1, ..., k_N\}$ be the set of critical scales corresponding to $\{v_1,...,v_N\}$ and 
     let $k_1'\leq...\leq k_N'$ denote a sorted sequence of such scales. 
    By definition of the critical scale, for any $k_i'$ in the sequence there are exactly $i$ samples falling within,
    and $N-i$ falling outside their bounds scaled by $k_i$, i.e., 
    \begin{eqnarray}
        y_j\in[\hat{y}_j-k_i\hat{z}_j^l, \hat{y}_j+k_i\hat{z}_j^u] \qquad\forall j:k_j'\leq k_i'\nonumber\\
        y_j\notin[\hat{y}_j-k_i\hat{z}_j^l, \hat{y}_j+k_i\hat{z}_j^u] \qquad\forall j:k_j'> k_i'\nonumber
    \end{eqnarray}
    (Note that in the case of ties only the element with highest index $i$ among the tie set is considered.)
    Thus, the fraction $\frac{N-i}{N}$ corresponds to the empirical miss rate as a function of $k$ 
    (see Eq. (\ref{Eq:MetricShorthand})), which is an estimator of the miss rate probability $P(Y\notin[\hat{Y}-k_i\hat{Z}^l, \hat{Y}+k_i\hat{Z}^u])$. 
    On the other hand,
    considering $K$ a critical scale of a randomly drawn sample, $V$, the fraction $\frac{i}{N}$ is an estimator 
    for the cumulative distribution function $P_K(k_i):=P(K\leq k_i)$. Hence
    \begin{equation}
        1-P_K(k_i)\equiv P(Y\notin[\hat{Y}-k_i\hat{Z}^l, \hat{Y}+k_i\hat{Z}^u]).\label{Eq:KToMissrate}\nonumber
    \end{equation}
\end{proof}

\subsubsection{Proof of Proposition \ref{Prop:missrate}}
\begin{remark}
    The bandwidth $\hat{\beta}$ and excess $\hat{\xi}$ (Eq. (\ref{Eq:MetricShorthand}))
    are monotonically increasing functions of the scale $k$.
\end{remark}

\begin{proof}
    Using the fact that $B\geq 0$, its expected value can be written as follows:
    \begin{equation}
        \left<B\right>_p = \int_{0}^{\infty} b p_B(b) db = \int_{0}^{\infty}[1-P_B(b)]db
        \label{Eq:BandwidthExpectedValue}
    \end{equation}
    where 
    $P_B$ 
    denotes the cumulative distribution function of $B$. 
    The second equality uses the tail expectation formula \cite{CasellaBook}.
    
    Since $1-P_B(b) = P(B>b)$ and $\beta$ is a monotonic function of $k$
    it holds that
    \begin{equation}
    P(K>k)\equiv P(B>b).
    \end{equation}
    From the above and the Lemma \ref{Lemma:1}, it follows that $P(B>b)$ corresponds to the miss rate 
    associated with the bandwidth $b=\beta(k)$:
    \begin{equation}
        P_m(b):=P(Y\notin[\hat{Y}-k\hat{Z}^l, \hat{Y}+k\hat{Z}^u])\equiv 1-P_B(b)
    \end{equation}
     Hence, Eq. (\ref{Eq:BandwidthExpectedValue}) becomes
    \begin{equation}
        \left<B\right>_{p_B} = \int_0^\infty P_m(b)db.\label{Eq:EBintegral2}
    \end{equation}
    Given $N$ samples, $\mathbf{v}=\{v_1, ..., v_N\}$ from $p_V$, we calculate the set of critical values, $\{k_1,..., k_N\}$.
    The sorted sequence, $k_1'\leq k_2'\leq...\leq k_N'$ gives rise to a sequence of bandwidths $b_1\leq...\leq b_N$. 
    The Riemann sum corresponding to the integral (\ref{Eq:EBintegral2}) is as follows
    \begin{equation}
        S(N) = \sum_{i=1}^N P_m(b_i') \Delta b_i\label{Eq:RiemannSum}
    \end{equation}
    with a partitioning determined by the sorted observations, $b_1\leq...\leq b_N$, $\Delta b_i = b_i-b_{i-1}$, $b_0=0$, and $b_i'\in[b_{i-1}, b_i]$. Choosing $b_i'=b_i$ we rewrite the sum (\ref{Eq:RiemannSum}) as
    \begin{equation}
        S(N) = \sum_{i=1}^N \hat{\rho}(k_i') [b(k_i')-b(k_{i-1}')]\label{Eq:RiemannSum2}
    \end{equation}
    with $\hat{\rho}$ being the empirical miss rate, as per Eq. (\ref{Eq:MetricShorthand}).
    
    Eq. (\ref{Eq:RiemannSum2}) corresponds to evaluating the area under the UCC using the rectangular rule. 
    The sum will approach the expected bandwidth value in Eq. (\ref{Eq:BandwidthExpectedValue}) as $\lim_{N\rightarrow\infty} S(N)$. Thus, the empirical AUUCC is an estimator for the expected bandwidth when using bandwidth-miss rate coordinates. 
    
\end{proof}
According to Proposition \ref{Prop:missrate}, given a dataset, and given the miss rate being one of the coordinates, 
the AUUCC amounts to the other metric's average over the entire operating range. A smaller AUUCC
relates to smaller average bandwidth (or excess) measurements as the calibration scale $k$ varies, as expected from prediction intervals of higher quality. 

\begin{corollary}
\label{Corollary1}
    The area under the UCC with excess-miss rate coordinates 
    is an estimator of the expected value $\left<X\right>_{p_X}$ with
    $X$ the excess random variable and $p_X$ its density function. 
\end{corollary}

The proof of Corollary \ref{Corollary1} follows trivially from the proof of Proposition \ref{Prop:missrate} by replacing
the bandwidth variable, $B$, with excess, $X$.

\subsubsection{AUUCC on the Excess-Deficit Coordinates}
\begin{proposition}\label{Prop:ExDef}
    Let $X$ and $D$ be the excess and deficit random variables generated by randomly selecting 
    a sample, $v$, determining its critical scale, $k$, and obtaining their values via Eq. (\ref{Eq:MetricShorthand}).
    Let the UCC be defined on the excess-deficit coordinates, $(\hat{\xi}, \hat{\delta})$, and the metrics 
    $\rho, \xi, \delta$ be differentiable and invertible functions. 
    The area under the UCC is an estimator of a quantity proportional to the expected value $\left<D\right>_q$ 
    with respect to a density given by $q(d)=\frac{p_D(d)}{p_X(\delta^{-1}(d))}/Q$, where $p_D$, $p_X$ denote deficit and excess
    densities and $Q$ is a normalizing constant, $Q=\int_0^\infty \frac{p_D(d)}{p_X(\delta^{-1}(d))}dd$. 
\end{proposition}
In this case, the interpretation involves an expectation of the deficit metric
proportional to a density {\em ratio} of the deficit and the excess. 

\begin{proof}
    Let the excess variable, $x=X$, be associated with the abscissa, 
    and $\delta(x)$ be the deficit function of the excess on the ordinate axis. 
    The AUUCC is 
    \begin{equation}
        \int_0^\infty \delta(x) dx.\label{Eq:EDAUC}
    \end{equation}
    Now consider the UCC a parametric curve parametrized by the miss rate, $r\in[0,1]$. 
    Let $r=\rho_X(x)$ and $r = \rho_D(d)$ where $\rho_{X,D}$ denotes a miss rate function of the excess and deficit, respectively. 
    Then $x=\rho_X^{-1}(r)$ and $d = \rho_D^{-1}(r)$, and Eq. (\ref{Eq:EDAUC}) can be rewritten as 
    \begin{equation}
        \int_1^0 \delta(\rho_X^{-1}(r))[\rho_X^{-1}]'(r)dr \label{Eq:EDAUC2}
    \end{equation}
    It is easy to show that $\frac{d}{dr}\left[\rho_X^{-1}(r)\right]=-\frac{1}{p_X(x)}$, where $p_X>0$ refers to the excess density. Hence (\ref{Eq:EDAUC2}) becomes
    \begin{equation}
        \int_0^1 \delta(\rho_X^{-1}(r)) \frac{1}{p_X(\rho_X^{-1}(r))} dr \label{Eq:EDAUC3}.
    \end{equation}
    After applying a variable change $r=\rho_D(d)$, Eq. (\ref{Eq:EDAUC3}) becomes 
    \begin{equation}
        \int_0^{\infty} d\cdot\frac{p_D(d)}{p_X(\delta^{-1}(d))}dd\label{Eq:EDAUC4}
    \end{equation}
    where $p_D$ refers to the deficit density. 
    We normalize the density ratio in Eq. (\ref{Eq:EDAUC4}) to obtain 
    \begin{equation}
        AUC = Q\cdot\int_0^{\infty}d\cdot q(d)dd\enspace \propto \left<D\right>_q \label{Eq:EDAUC5}
    \end{equation}
    whereby $q(d):=\frac{p_D(d)}{p_X(\delta^{-1}(d))}/Q$ and $Q=\int_0^{\infty}\frac{p_D(d)}{p_X(\delta^{-1}(d))}dd$.
    Thus, the Eq. (\ref{Eq:EDAUC5}) shows the AUUCC is proportional to the expected deficit with respect to the  distribution, $q$.
    Using the Riemann sum argument, similar to one in the proof to Proposition \ref{Prop:missrate}, 
    it is straight-forward to show that the empirical AUUCC is an estimator for 
    (\ref{Eq:EDAUC5}) up to the constant $Q$.
\end{proof}

In the case of Proposition \ref{Prop:ExDef}, the interpretation involves again an expectation of one of the axes' metrics, namely the deficit, however,
with respect to a distribution of a density ratio between the deficit and the excess. Similar to the previous
result, a smaller AUUCC relates to a smaller deficit average with respect to the density, $q$.
One example of such average being small would be a case where the mode of $p_D$ lies near zero deficit and 
the corresponding $p_X$ is small there, with its mode residing at higher deficits, thus concentrating the mass of $q$ around small deficit values.

Exploiting these results in the {\em optimization} of models to produce better prediction intervals appears an interesting avenue for future work.

\subsection{Cost Function}
\begin{figure}[hbt]
  \centering
  \includegraphics[height=3.4cm, width=5cm]{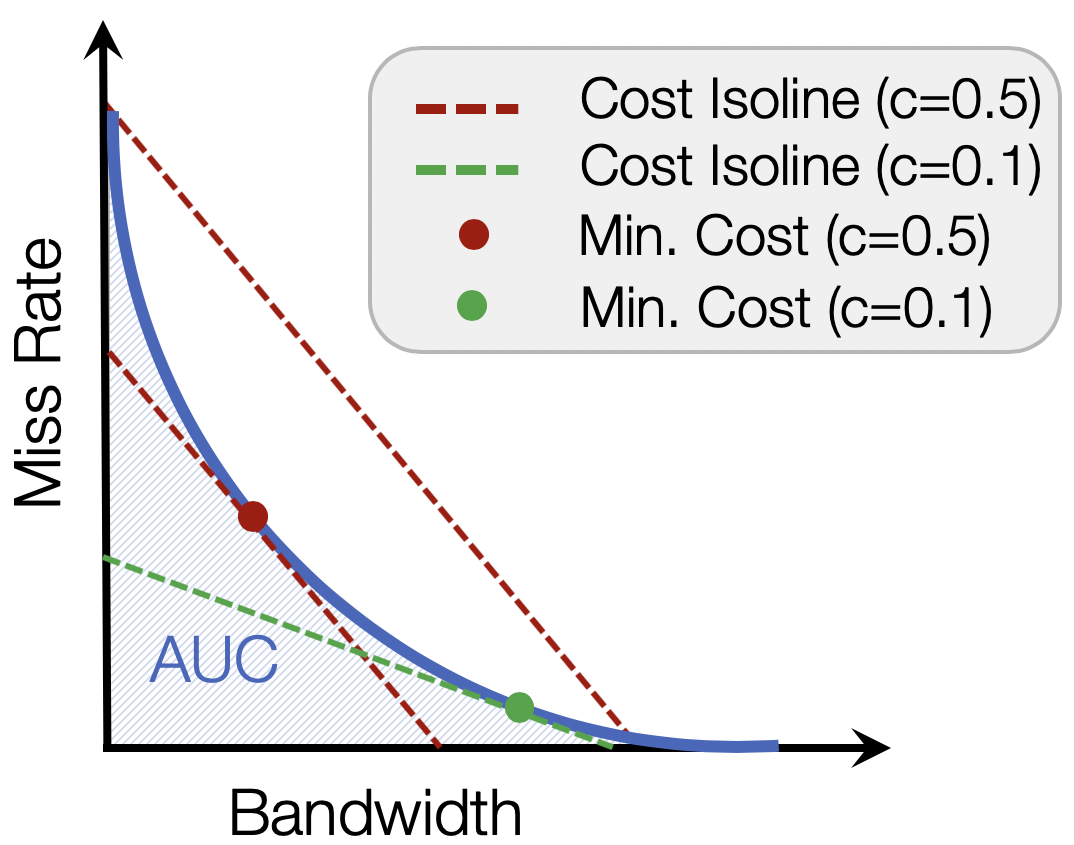}
  \captionof{figure}{Illustration of the Area Under Curve (AUC) and Cost within the UCC graph }
  \label{Fig:UCCExample2}
\end{figure}

Considering the cost trade-off between the two axes at a particular operating point, it is useful to define a function combining the two in a meaningful way. The simplest example is a linear cost function
\begin{equation}\label{Eq:CostFunction}
    C(k) = c \hat{\beta}(k) + (1-c) \hat{\rho}(k), \enspace c\in[0,1]
\end{equation}
that uses an application-dependent factor, $c$, to focus on a specific area of the operating range (e.g., low miss rate area). On the UCC coordinate system, $C(k)=const$ shows as an isocost line (see Figure \ref{Fig:UCCExample2}) whose slope is proportional to $-c/(1-c)$. A minimum achievable cost, $C(k^*)$ with $k^*=\argmin_k C(k)$, is an intersection of a model's UCC and the minimal isocost as illustrated in the example. In this context, the UCC provides for a visual assessment between the original OP cost and the optimum cost as well as gives the scaling $k^*$ needed to reach that optimum. 

\begin{remark}\label{Rem:MAE}
    Let $d_i=|\hat{y}_i - y_i|$. Given a scale $k$, and symmetric prediction bands $\hat{z}_i:=\hat{z}_i^l = \hat{z}_i^u$, the linear cost (\ref{Eq:CostFunction}) with $c=0.5$ at any operating point $k$ on the excess-deficit coordinate system corresponds to half of the mean absolute error (MAE) between the
    absolute difference and the scaled band:
        $MAE(k) = \frac{1}{N}\sum_i |d_i-k\cdot z_i|$.
\end{remark}
\begin{remark}\label{Rem:IntervalScore}
    For the choice of $f_1=\hat{\beta}, f_2=\hat{\delta}$ and $c=\frac{1}{\alpha+1}$ with $\alpha\in[0,1]$ denoting the confidence level, the symmetric cost (\ref{Eq:CostFunction}) corresponds to the well-known Interval Score (see \cite{Gneiting2007}, Section 6.2), up to a scale $\frac{\alpha+1}{\alpha}$.
    
\end{remark}

\end{document}